%% file: main_Tom_AICS.tex
\documentclass[letterpaper]{article} % DO NOT CHANGE THIS
\usepackage{aaai24}  % DO NOT CHANGE THIS
\usepackage{times}  % DO NOT CHANGE THIS
\usepackage{helvet}  % DO NOT CHANGE THIS
\usepackage{courier}  % DO NOT CHANGE THIS
\usepackage[hyphens]{url}  % DO NOT CHANGE THIS
\usepackage{graphicx} % DO NOT CHANGE THIS
\usepackage{natbib}  % DO NOT CHANGE THIS AND DO NOT ADD ANY OPTIONS TO IT
\usepackage{caption} % DO NOT CHANGE THIS AND DO NOT ADD ANY OPTIONS TO IT
\usepackage{algorithm}
\usepackage{algorithmic}

\usepackage{newfloat}
\usepackage{listings}
\DeclareCaptionStyle{ruled}{labelfont=normalfont,labelsep=colon,strut=off} % DO NOT CHANGE THIS
\floatstyle{ruled}
\newfloat{listing}{tb}{lst}{}
\floatname{listing}{Listing}
\usepackage{amsmath}
\usepackage{amssymb}
\usepackage{booktabs}
\usepackage[export]{adjustbox}

\usepackage{scalerel}
\usepackage{longtable}
\usepackage{subcaption}
\usepackage{xcolor}
\usepackage{soul}
\usepackage{color}
\usepackage{siunitx}
\usepackage{arydshln}

\usepackage{multirow}
\usepackage{tabularx}

\usepackage{microtype}
\usepackage{soul}
\usepackage[utf8]{inputenc}
\usepackage[T1]{fontenc}

\title{A Generative Adversarial Attack for Multilingual Text Classifiers}
\author {
    Tom Roth\textsuperscript{\rm 1,\rm 2},
    Inigo Jauregi Unanue\textsuperscript{\rm 1,\rm 3},
    Alsharif Abuadbba\textsuperscript{\rm 2},
    Massimo Piccardi\textsuperscript{\rm 1}
}
\affiliations {
    \textsuperscript{\rm 1}University of Technology Sydney, NSW, Australia\\
    \textsuperscript{\rm 2}CSIRO's Data61, Sydney, Australia\\
    \textsuperscript{\rm 3}RoZetta Technology, Sydney, Australia\\
}

\begin{document}

\maketitle

\begin{abstract}
Current adversarial attack algorithms, where an adversary changes a text to fool a victim model, have been repeatedly shown to be effective against text classifiers. These attacks, however, generally assume that the victim model is monolingual and cannot be used to target multilingual victim models, a significant limitation given the increased use of these models. For this reason, in this work we propose an approach to fine-tune a multilingual paraphrase model with an adversarial objective so that it becomes able to generate effective adversarial examples against multilingual classifiers. The training objective incorporates a set of pre-trained models to ensure text quality and language consistency of the generated text. In addition, all the models are suitably connected to the generator by vocabulary-mapping matrices, allowing for full end-to-end differentiability of the overall training pipeline. The experimental validation over two multilingual datasets and five languages has shown the effectiveness of the proposed approach compared to existing baselines, particularly in terms of query efficiency. We also provide a detailed analysis of the generated attacks and discuss limitations and opportunities for future research. 
\end{abstract}

\section{Introduction}
\label{sec:introduction} 
Advancements in machine learning have led to models that achieve remarkable performance across a variety of natural language processing (NLP) tasks. Despite this progress, these models remain susceptible to adversarial attacks, which manipulate inputs to induce incorrect predictions while maintaining linguistic coherence and fluency. In this work we consider text classification, for which a wide variety of adversarial attacks have been developed, ranging from simple character or token replacements to fully generative models. 

Text classification models can be developed to process only a specific language (i.e., they are monolingual) or to process multiple languages (i.e., they are multilingual). 
As processing multiple languages with a single model is more scalable and efficient than training a number of monolingual models, multilingual models are now widely used across many NLP applications. Despite this, multilingual models have been largely overlooked as victims of adversarial attacks, with the majority of existing attack algorithms being implicitly monolingual and ill-suited to handle multiple languages.

%are typically developed to operate only for a specific language (i.e., they are monolingual) \textcolor{red}{Inigo: Not sure if this statament is too broad, nowadays it is common to have multilingual text classifiers. Maybe we can change the wording to ``models can be developed to operate only for a specific language (i.e., monolingual) or to process multi
The most common approach for creating text adversarial examples is to leverage  combinatorial optimisation in some form. In this approach, an attack takes in input a correctly-classified example and attempts to transform it into an adversarial one  by iteratively modifying its individual tokens. The modifications adhere to a set of allowed transformations, such as word insertion or replacement, and are guided by a chosen objective function and search method. Constraints are also imposed to ensure the validity of the transformations, such as requiring semantic similarity to the original input. The process typically stops once a successful adversarial example is found, or the maximum number of attempts is reached. De facto, these attacks solve a separate optimisation problem for each example and are highly effective --- provided that they are allowed enough steps, time and computational resources. However, they are typically slow and hard to scale, and the space of the adversarial examples is constrained by the choice of the allowed transformations, which may limit their richness and diversity.

An alternative approach is to use a trainable generative model to produce the adversarial examples. These models learn a mapping from an original example as input to an adversarial example as output based on a training dataset. At inference time, they are able to produce an adversarial attack from the given input in a single forward pass. Since these models do not perform an optimisation process for each input example, they are, in principle, unlikely to match the performance of the transformation-based approach. However, the generative approach mitigates many of its disadvantages: 1) producing the adversarial example is much faster, requiring only a single forward pass per batch of examples; 2) multiple adversarial candidates can be generated simultaneously just by using beam decoding; and 3) it can potentially perform richer and more diverse transformations of the input example.

With this motivation, in this paper we propose an approach for training a generative model to attack a multilingual victim model, under a white-box assumption\footnote{Which holds for all publicly-distributed models, such as those on Hugging Face and GitHub.}. We begin with a generative pre-trained multilingual model (specifically, mT5 \citep{mT5}) capable of generating text across more than a hundred languages. Although pre-trained, this model does not yet solve any task, and so we first train it with a multilingual paraphrasing objective. This allows the model to first learn a diverse range of text-to-text transforms in multiple languages. As second stage of training, we fine-tune the generative model with an adversarial training objective. This training objective incorporates both the victim model itself, and also a number of other pre-trained models that help maintain the linguistic properties of the text. Amongst them, a language detection model prevents the generative model from outputting text in languages different from the original's, which is an implicit requirement of this task and a key contribution of the proposed approach. The models are connected with the generative model by matrices that remap their vocabularies, allowing the use of any models in the training objective, while at the same time retaining the end-to-end differentiability of the pipeline.  At inference time, the model receives an input in language $L$ and generates adversarial candidates in the same language, provided $L$ is amongst the languages it was fine-tuned in.

To comparatively evaluate the performance of the proposed approach, we have adapted two existing monolingual optimisation attack algorithms to the multilingual case, and used them as baselines. As datasets, we have used two multilingual datasets that incorporate four and five languages, respectively. The experimental results show that our generative model is a fast, diverse and effective attacker, and matches the performance of baselines that run for far longer. In summary, our paper makes the following main contributions:

\begin{enumerate}
\item proposes a generative attack for multilingual victim models (a first to the best of our knowledge); 

\item proposes a dedicated training objective that leverages the victim model itself and various other modules that promote desirable linguistic properties;

\item makes use of vocabulary-mapping matrices to allow maximum flexibility in the choice of the models and ensure differentiability end-to-end; 

% \item we extend two existing optimisation attacks to the multilingual scenario 

\item lastly, our experimental results over two multilingual datasets and five languages show a remarkable comparative performance against the chosen baselines.
\end{enumerate}

\section{Related Work}
\label{sec:related}
\textbf{Combinatorial optimisation attacks.} Combinatorial optimisation attacks are undoubtedly the most common type of attack for text classifiers. Many attacks have been proposed, each defined by their own allowed transformations, search method, and constraints. Well-known attacks include TextFooler \citep{TextFooler}, BERTAttack \citep{li-2020-bert-attack}, BAE \citep{BAE} and CLARE \citep{CLARE}. The most recent research has focused on a range of aspects, such as increasing the readability of the adversarial examples \citep{TextGuise}, improving the query efficiency of the attacks \citep{BeamAttack, QueryEfficient} and adapting adversarial attacks designed for images to the text domain \citep{BridgeTheGap}. 

\textbf{Generative attacks.} Although less explored, a number of generative approaches have been proposed. The most relevant to our work are the variations on the white-box, differentiable model-cascading design described in the Introduction. Such work includes \citet{CATGen}, which has incorporated a downstream model to control the topic of the generated adversarial candidates at inference time, and \citet{greybox}, which has incorporated several features (e.g., label smoothing and copy mechanisms) to improve the quality of the generated examples. In turn, \citet{Guo2021Gradient} have used pre-trained models to learn an example-specific matrix of token probabilities, which can be repeatedly sampled to generate adversarial examples. To avoid the use of vocabulary-mapping matrices (or similar) previous work has largely assumed a scenario where all the component models have to share the same vocabulary, which is a very restrictive assumption. An exception is \citet{Song2021}, which has constrained the generative model to only output common tokens to all vocabularies. However, this is also very restrictive. 
 %Models have included GANs, LSTMs, variational autoencoders, and transformers (CITE FOR ALL). 
 % However, these existing methods have not introduced a systematic and flexible way to train generative models that can also satisfy all the desired adversarial attack properties.  

\textbf{Multilingual attacks.} Most existing text adversarial attacks are monolingual, and predominantly in English. Some work has partially addressed the second restriction by designing attacks for other languages, such as Chinese \citep{ExpandingScope, SemAttack} or Arabic \citep{Alshemali2019}. The only multilingual attack we are aware of is \citet{Rosenthal2021AreMB}, who have attacked a multilingual question answering system by adding distracting statements to the question. Other multilingual attacks have been developed for the related task of \textit{code-mixing}, where a single input contains multiple languages (mimicking how multilingual speakers sometimes mix multiple languages when speaking). 
% These attacks, too, target multilingual victim models. 
Code-mixing attacks apply various perturbations to the example: \citet{CodeMixing} have introduced words and phrases through the use of bilingual dictionaries and translation models, and \citet{Das2021AdvCodeMixAA} have made phonetically-similar replacements for the Bengali-English and Hindi-English scenarios. However, the code-mixing scenario is different from what we target in our paper: in our case, each example in input to the multilingual classifier is assumed to be in only one language. This allows enforcing a language consistency constraint as a key requirement for the acceptability of the attack.

% Text adversarial attacks operate under varying threat models and assumptions, broadly categorised into white-box and black-box scenarios. White-box attacks assume comprehensive access to the model, including its architecture and weights, while black-box attacks are restricted to querying for class predictions. The prevailing methodology for these attacks is combinatorial optimisation. Token transformations in attacks span a range of granularities, from character-level manipulations as in HotFlip, to subword alterations in BertAttack, and up to word and phrase level modifications in TextFooler and respective phrase-level studies. These attacks are primarily token-modification in nature, encompassing a goal function, allowed transformations, constraints, and a search method.

% White-box token-based attacks date back to at least the work of \citet{Papernot2016}. Typically, these attacks leverage the gradient signal of the victim model in two main ways. The first is to rank token importance in the original sentence, thus identifying promising attack targets, as demonstrated in \citet{Wallace2019AllenNLP}. The second is to aid in selecting token transformations that best meet adversarial criteria, as shown in various character-level and word-level attacks \citep{HotFlip, MHA, HotPhrases}.

\section{Proposed Approach}
\label{sec:proposed_approach}

\begin{figure} [!ht]
  \centering
    \includegraphics[width=\columnwidth]{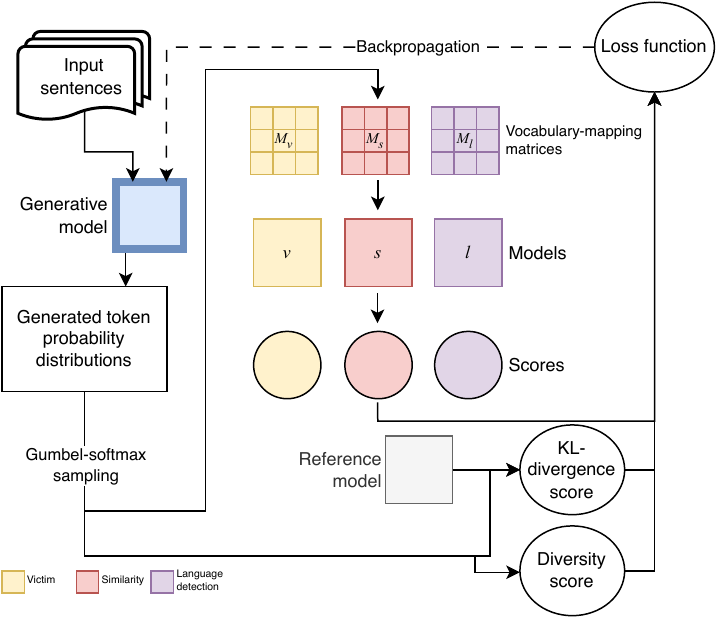}
  \caption{The proposed approach. The generative model has been trained using a loss function comprised of a number of factors, including scores from three component models (victim, similarity, and language detection), a KL divergence score, and a diversity score.}
  \label{fig:experimental_setup}
\end{figure}

The training approach fine-tunes a pre-trained generative model $g$ so that it can generate adversarial examples for a victim model $v$. Besides the victim, the approach incorporates two additional component models into the training objective: a semantic similarity model $s$ and a language detector model $l$. All models involved have an associated tokeniser with a specific vocabulary, which for model $i$ is denoted by $V_i$. The parameters of all models are fixed during training, except for $g$. After fine-tuning, $g$ is able to generate adversarial examples for $v$. The  training setup is shown in Figure \ref{fig:experimental_setup}. 

% We aim to fine-tune a generative model $g$, with parameters $\theta$ and vocabulary $V_g$, to generate adversarial examples for victim model $v$, with vocabulary $V_v$. The approach includes two additional component models for the training objective: a semantic similarity model, $s$, of vocabulary $V_s$, and a natural language inference (NLI) model, $n$, of vocabulary $V_n$. The parameters of models $v$, $s$ and $n$ are all fixed, while those of $g$ are the target of the proposed training approach. 

\subsection{Pre-Training}
For our generative model, we start from a pre-trained mT5-base \citep{mT5} checkpoint. This model has been trained using a span-corruption objective, and has not been fine-tuned on any downstream task, so the first pre-training step we take is to fine-tune it on a multilingual text paraphrasing training dataset. After this training step, the model is able to generate a range of diverse paraphrases across multiple languages. This gives the starting weights for the generative model $g$, that is later fine-tuned. We also take a copy of this model and fix its weights to serve as the reference model $g^*$ (see Section \ref{sec:loss_function}).  

\subsection{Fine-Tuning}
During the forward pass, the input is an original example $x$ which is passed to $g$, generating output $x'$ with length $T$. The corresponding sequence of token probability distributions forms a matrix $P$, with dimensions $T \times |V_g|$.

The generated text cannot be passed to the objective models as is, as that would involve non-differentiable operators (either sampling or argmax). Therefore, we employ a different approach to retain the differentiability of the pipeline: passing a weighted average of token embeddings to the downstream models. This is created by combining $P$ with the token embedding matrix of the downstream model, and a second matrix that maps the vocabulary of the generative model to the vocabulary of the downstream model. We call this second matrix a \textit{vocabulary-mapping matrix} and describe it in detail in Section \ref{sec:vocabulary-mapping}. 

% \footnote{  Black-box generative attacks, not using gradient information from the victim model, are free to pass the generated text between the models.}
More formally, for any component model $V_i$, with  $i \in \{v, s, l\}$, we have computed the respective weighted embeddings $W_i$ as: $$W_i = P M_i E_i $$ where $E_{i}$ is the token embedding matrix of model $i$, and $M_i$ is the vocabulary-mapping matrix that maps $V_g$, the vocabulary of $P$, to $V_i$, the vocabulary of model $i$.

As done in prior research \citep{greybox, CATGen}, we have used the Gumbel-softmax reparametrisation trick \citep{GumbelSoftmax} and replaced $P$ with a sampled matrix $P_b$ that has incorporated $\text{Gumbel}(\tau)$ noise. Here, $\tau$ is a temperature parameter that controls the entropy of the resulting probability distributions. This step introduces stochasticity into the token distribution with the aim to increase exploration during training. %Values of $\tau > 1$ make the samples more evenly distributed, while values $< 1$ concentrate them towards a one-hot distribution.  This step 

Afterwards, the weighted embeddings $W_i, i \in \{v, s, l\}$ are passed to their respective downstream models, and the output from each downstream model is used in the loss function (Section \ref{sec:loss_function}). The backward pass computes the chained gradients and updates the parameters of the generative model, using standard backpropagation.

\subsection{Vocabulary-Mapping Matrices}
\label{sec:vocabulary-mapping}
Although required for our differentiable training approach, it is not straightforward to map tokens between vocabularies, particularly when they are of different sizes and have been constructed with different tokenisation algorithms.
In our implementation, the vocabulary of the generative paraphrase model is constructed using the SentencePiece \citep{SentencePiece} tokenisation algorithm, while the downstream models' vocabularies use WordPiece \citep{WordPiece}, and differ in content and size. We have been able to construct a workable mapping with the following rules:
% In our implementation, the generative model's tokeniser uses SentencePiece \citep{SentencePiece}, while the tokenisers of all the component models use WordPiece \citep{WordPiece}. We match tokens where possible using string matching rules, and use the component model tokeniser to create mappings for the remainder.
% \section{Token mapping rules}
% \label{sec:token_mapping_rules}

\begin{enumerate}
    \item Map one-to-one all direct matches between SentencePiece start-of-word tokens and WordPiece non-continuation tokens.
    \item Map one-to-one all direct matches between SentencePiece non start-of-word tokens and  WordPiece continuation tokens.
    \item Map any special tokens (e.g., PAD, EOS, UNK) directly across both vocabularies. Map the extra-id tokens in the generator's vocabulary to the UNK WordPiece token.
    \item Map the remaining SentencePiece tokens one-to-many with WordPiece tokens using the WordPiece tokeniser, stripping any generated special tokens, and assigning equal probabilities to all matches. Map any remaining tokens (e.g. special cases like \textbackslash xad)  to the UNK token.
\end{enumerate}
The end result is a matrix $M_i$ that maps tokens from the vocabulary of the generative model, $V_g$, to $V_i$, the vocabulary of the downstream model. In $M_i$, each row is a probability distribution that represents the one-to-many token mapping, with the sum of each row being 1, and the vast majority of entries being 0. The shape of $M_i$ is $|V_g| \times |V_i|$, and as this is a large matrix\footnote{$|V_g|$ alone is 250K.}, to save space we store it as a sparse matrix.

\subsection{Loss Function}
\label{sec:loss_function}
The training objective, denoted as $o(x, x')$, has been designed to balance the adversarial and linguistic factors required by a successful multilingual adversarial attack. It is given by %Our aim is to create adversarial examples that successfully flip the predicted labels, but also retain the text's original meaning and keep the language consistent to the original.  
% Our loss function balances balancing each component to ensure adversarial examples that alter predicted labels while preserving original meaning and linguistic acceptability, as per the definition of \citep{michel-etal-2019-evaluation}.

% Our approach, in line with the explanation from \citep{michel-etal-2019-evaluation}, is to employ a multi-objective loss function that delicately balances each component. This method yields adversarial examples that alter predicted labels without compromising the original meaning and linguistic acceptability.

% In accordance with the definition provided by \citet{michel-etal-2019-evaluation},

\begin{equation}
\label{eqn:training_objective}
\begin{split}
o(x, x')  =  \alpha_v t(v(x,x'), \beta_v) + \\ \alpha_s t(s(x,x'), \beta_s)  + \\ \alpha_l t^*(l(x,x'),\beta_l) - \\ \alpha_{KL} t^*(D_{KL}(x,x'), \beta_{KL})
\end{split}
\end{equation}

\noindent where: 

\begin{itemize}

\item $v(x,x')$ is the `victim model score': the degradation in victim model confidence of the true class when replacing $x$ with $x'$.
%It measures how much the classifier's confidence in the correct class drops when replacing $x$ with $x'$ 
% repr$ = f(x)_y - f(x')_y$ is the \textit{victim model score} where $f$ is the classifier, $x$ the original with class $y$, and $x'$ the paraphrase. This term measures the degradation in victim model confidence of the true class.  

\item $s(x,x')$ is the `similarity score' between $x$ and $x'$, given by the cosine similarity of their sentence embeddings from a pre-trained multilingual Sentence-BERT model \citep{reimers-2019-sentence-bert}.  %which measures similarity in semantic content between original and paraphrase.  We extract sentence embeddings of both using a pretrained Siamese-BERT model \citep{reimers-2019-sentence-bert} and compute the cosine similarity. 
\item  $l(x,x')$ is the `language consistency score', given by the degradation in original-language confidence between $x$ and $x'$, using  a pre-trained language detection model.  %that $x'$ is the same  use a pre-trained language detection model to classify the language of $x$ and $x'$, and set this to the difference in confidence of the ground-truth language 

\item $D_{KL}$ is a Kullback-Leibler (KL) divergence term, calculated between the token probabilities produced by the fine-tuned generative model $g$ and a reference model $g^*$. This term prevents the fine-tuned distribution from deviating too much from the reference, which we take as the multilingual paraphrase model before fine-tuning, allowing the beneficial paraphrasing qualities of \( g^* \) to be maintained.\footnote{Due to memory constraints, $g^*$ is only a part of the initial paraphrase model.} $D_{KL}$ is defined as:
\begin{equation}
\label{eq:kl}
\begin{aligned}
D_{KL} =  \frac{1}{T}  \mathbb{E}_{x \sim \mathcal{D}, x' \sim g(x;\theta)} 
[\log p_g(x'|x) - \\ \log p_{g^{*}}(x'|x) ]
\end{aligned}
\end{equation}
In (\ref{eq:kl}), the divergence has been normalised by the generated sequence length, $T$, to prevent longer sequences from being unfairly penalised. 

\item $t$ and $t^*$ are threshold clipping operators, with $t(a,\beta) = a$ if $a < \beta$, and $0$ otherwise, and $t^*(a,\beta) = a$ if $a > \beta$, and 0 otherwise. These operators are intended to  encourage a balanced optimisation of the multiple objective components,
% , and to prevent over-optimisation of a single component.
with hyperparameters $\alpha$ and $\beta$ controlling the contribution of each component.
%\textcolor{red}{Is it worth reiterating which are those objective components? (i.e., label flip, semantic similarity and language preservation)}
\end{itemize}

This training objective $o(x, x')$ is then incorporated into a \textit{batch-level} loss, $L$, which is defined as: 

% \begin{equation}
% \label{eqn:loss_fn}
% \begin{split}
% L = - \mathbb{E}_{(x,x' \in B)} l(x,x')  + \alpha_{div} D_{div} (B)
% \end{split}
% \end{equation}

\begin{equation}
\label{eqn:loss_fn}
\begin{split}
L = - \left(\frac{1}{|B|}\sum_{(x,x')\in B}o(x,x')\right)  + \alpha_{d} \hspace{2pt} d(B)
\end{split}
\end{equation}

\noindent where $d(B)$ is a batch-level diversity score, and $\alpha_{d}$ its corresponding coefficient. 

To compute $d(B)$, within batch $B$ and using the same model as for the similarity score, we calculate two cosine similarity matrices: one of the mean of the token embeddings for each generated example, and one of the embeddings of the original input text.  We then take the mean of the squared difference of the upper triangular matrices as the diversity penalty $d(B)$. This term encourages the generated examples to maintain a similar batch diversity as for the original examples, as to prevent degeneration during training and improve the variety of the generated examples. 

%and then calculate gumbel samples computed as the mean of the pairwise cosine distance between the sentence embeddings of the examples in batch $B$, calculated with the same model as the similarity score, and . 

% \textcolor{red}{ Not sure if these eqns are better than the expectation one? M+I to check}
% \begin{equation}
% \label{eqn:loss_fn2}
% \begin{split}
% L  =  - \left[ \frac{1}{|B|} \sum_{(x,x' \in B)} l(x,x')  \right] + \gamma D_{div}
% \end{split}
% \end{equation}

% Finally, we apply a threshold clipping operation, defined as $h(x, t, m)$, to each term in $L$, excluding the diversity penalty. This operation aims to prevent over-optimisation of any one component. The function $h$ is defined as:

% where $x$ is a term element, $t$ the corresponding threshold, and $m$ controls the inequality direction (-1 for $x<t$, 1 for $x>t$). This process ensures balanced optimisation across components.

As each individual component in the loss function is differentiable and their combination is linear, the overall loss function is also differentiable and the system can be efficiently optimised using backpropagation. The overall design is  flexible, and each coefficient can be adjusted to prioritise different objectives, such as attack strength, textual fluency or language consistency.

% adversarial examples with particular properties. Increasing one coefficient implicitly makes the others less important. For example, increasing $\alpha_{flu}$ will generate more fluent adversarial examples, but on average will lower the attack success rate and semantic similarity, and increase the contradiction chance. 

% \textbf{Is fluent.} We are not interested in paraphrases that do not read like natural English. To assess this, we 

% \textbf{Retains the true label.} The original and paraphrase must have the same ground-truth label. We assess this by 

% \textbf{Is semantically consistent.} The original and paraphrase must have (broadly) the same semantic content. 

% eDesigning a loss function requires first a precise definition of an adversarial example.  We follow previous work \citep{michel-etal-2019-evaluation} and conclude that in addition to switching the predicted label, an adversarial example should both preserve meaning and be linguistically acceptable. We enforce these principles by incorporating the following terms into our loss function. 

\subsection{Validation and Early Stopping}
\label{sec:validation}
The quality of the generated text and its attack strength are somehow inversely proportional to the duration of training: shorter training yields high-quality text with lower attack strength, while prolonged training increases attack strength, yet degrades text quality. Deciding when to stop training is hence an important decision, and we have used the following stopping criteria: during the validation phase, we generate 16 candidates per original example, decoding using \textit{diverse beam search} \citep{Vijayakumar2016}. Each candidate is then given a `score' based on its adversarial and linguistic qualities: $$\gamma_v v(x,x') + Q(x,x'),$$ with $$Q(x,x') = \gamma_s s(x,x') - \gamma_l l(x,x') - \gamma_{KL} D_{KL}(x,x')$$ being a `text quality' score. The $\gamma$ coefficients set the balance between attack strength and text quality.

The validation metric we use is the average score across all candidates for the examples in the validation set.  We calculate the metric after every multiple of training steps, and halt the training process once the metric fails to improve for a number of steps, as standard in early stopping.

\section{Experimental Setup}

\subsection{Datasets and Models} 
\label{sec:datasets}
We have evaluated the approach on two multilingual datasets. The first is the Multilingual Amazon Reviews Corpus (MARC) \citep{MARC}, comprised of Amazon customer ratings of the goods they purchased. For our experiments, we have used the English, German, French, and Spanish language splits. The second dataset is the Tweet Sentiment Multilingual (TSM) dataset \citep{TSM}. This is a multilingual sentiment analysis dataset that categorises a variety of tweets as negative, neutral or positive. For this dataset, we have used the above four language splits, plus Arabic. Finally, to initially train the multilingual paraphrasing objective, we have used the TaPaCo dataset \citep{TaPaCo}, selecting the splits that correspond to all languages covered above. For all datasets, we have filtered out all examples longer than 32 tokens, as these longer examples are difficult to paraphrase effectively. 

As multilingual victim models, we have fine-tuned a pre-trained multilingual DistilBERT model \citep{DistilBERT} on the MARC and TSM datasets separately, using only the languages mentioned above. The performance of the resulting trained models is close to the state-of-the-art performance for both datasets, making the adversarial attacks relevant. Full details of datasets and models, including the specific choice of component models, are in Appendix \ref{sec:hyperparameters}.

\subsection{Baselines}
\label{sec:baselines}
As no general multilingual adversarial attack currently exists, for the baselines we have extended CLARE \citep{CLARE} and BAE-R \citep{BAE}, two existing monolingual combinatorial optimisation attacks, to the multilingual case. We term these new multilingual variants mCLARE and mBAE. Both attacks originally used a BERT masked language model \citep{BERT} to generate the transform token candidates, and the Universal Sentence Encoder \citep{UniversalSentenceEncoder} to assess sentence similarity. In their multilingual extension, we have replaced them with a multilingual BERT masked language model and the multilingual Universal Sentence Encoder \citep{MultilingualUSE}, respectively. We have also removed language-dependent constraints from the attacks that originally forbade changing stop-words and restricted attack targets to words with a certain part-of-speech tag. The final mCLARE attack is able to replace, insert, or merge individual tokens, while mBAE can only replace.

Since the attack performance for these baselines directly depends on the number of times that the attack is allowed to query the victim model, to get an accurate view of performance we have run the baselines a number of times, each with a differing amount of maximum queries. The number of queries can also be seen as a proxy for the runtime of the method, for which there is a clear correlation \citep{yoo2020searching}. However, the actual runtime directly depends on the time per query, which is approximately 10x larger for mCLARE than mBAE.

% restricting CLARE, we also remove a constraint which forbids changing stop-words, as their lists of stop-words are language dependent. 

% Since it is difficult to find baselines not in English (see \citep{sec:related}) we adapt two commonly used adversarial attacks, TextFooler \citep{Jin2020TextFooler} and BERTAttack \cite{li-2020-bert-attack}. These are both are token-replacement attacks that replace individual tokens sequentially in a constrained optimisation process.  The original TextFooler replaces words based on close points in word embedding space. While they use English word embeddings, we adapt it to 9jsteadn use multilingual word embeddings. We have multiple versions of this constraint. The first is where the attacker first detects the language, and then loads word embeddings in that language. The second is when the attacker uses multilingual word embeddings. 
% Similarly, we adapt BERTAttack to the multilingual space. BERTAttack, as it is named, uses BERT to select replacements for individual tokens. We instead use mBERT, in the multilingual setting, or load a language-specific pretrained BERT, in the monolingual case. 

\vspace{-6pt}

\subsection{Candidate Selection}
At inference time, our generative fine-tuned model is capable of generating, in principle, multiple adversarial candidates per input example by using beam search. However, for fairness we have opted to select only one candidate from our model, because it may be otherwise unfair to compare with the combinatorial optimisation attacks which return only one adversarial example per input. 
To select our candidate, we start by decoding using diverse beam search \citep{diverse_beam_search} to create $n$ outputs for each original example. Then, out of those that flip the label, we select that with the highest text quality score, $Q(x,x')$, as the candidate (see Section \ref{sec:validation} for details). 

\begin{figure*}[h]
\centering

\begin{subfigure}{\textwidth}
\includegraphics[width=\textwidth]{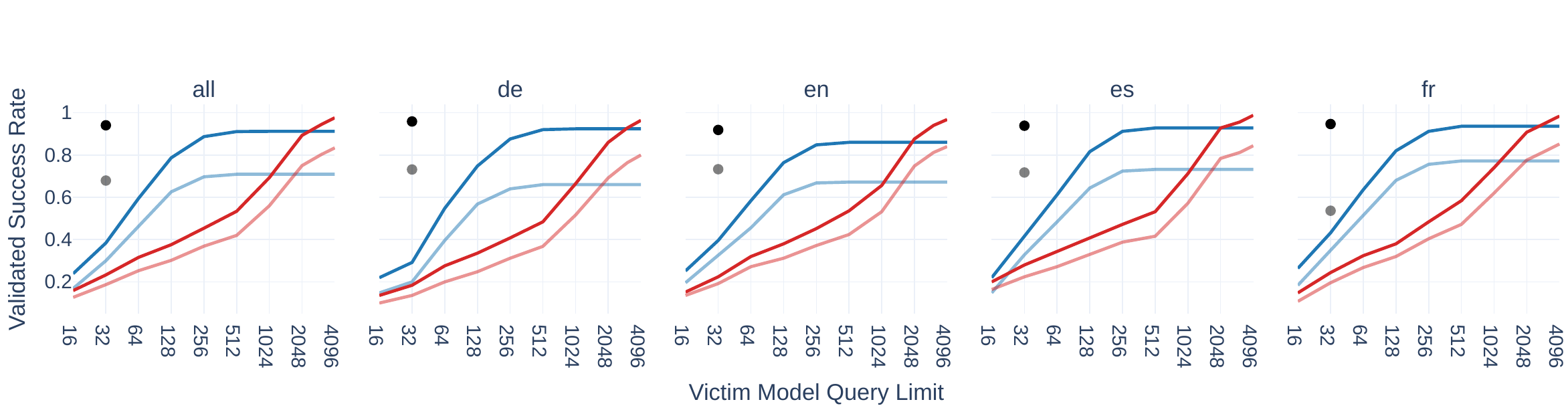}
\caption{Results for the MARC dataset. }
\label{fig:amazon}
\end{subfigure}

\vspace{12pt} % Add some space between the two subfigures

\begin{subfigure}{\textwidth}
\includegraphics[width=\textwidth]{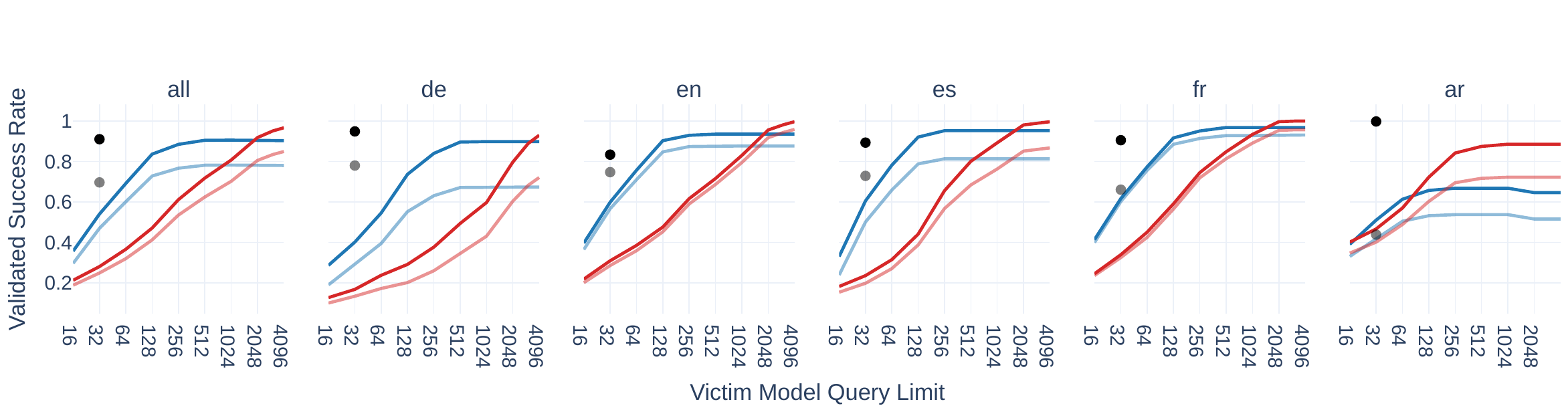}
\caption{Results for the TSM dataset. }
\label{fig:tweet}
\end{subfigure}

\caption{Results for the MARC dataset (above) and the TSM dataset (below), overall and split by language. The x-axis is the maximum number of victim model queries,
% (a proxy for the runtime)
and the y-axis is the validated success rate (VSR) of the attacks. For all methods, we report VSR values for two different fluency threshold (see Section \ref{sec:eval_metrics}). The black dots plot the values for the proposed approach (averaged across three random seeds), while the lines plot the values for the baseline methods, with blue for mBAE and red for mCLARE.
% requiring only one query per generated candidate (to check if it is successful), 
% As determining thresholds for the validated success rate is inherently subjective, we have explored two different fluency thresholds for this criterion: from darkest to lightest, the cutoffs are -11, and -9, with less negative numbers (i.e. towards zero) being more fluent.
The results show that the proposed approach has achieved remarkable VSR values in many cases, and has achieved an impressive trade-off between performance and number of queries. At least one of the  baselines eventually surpasses its performance in many cases, but only after many more queries (512, 1024 or more). }
\label{fig:combined-results}
\end{figure*}

\input{examples_table}

\input{ablation_sim_table}
\input{ablation_lang_table}

\subsection{Evaluation Metrics}
\label{sec:eval_metrics}

% problems with automated measurement - TO INCORPORATE 

% * bertscore limitations: https://aclanthology.org/2021.wmt-1.59.pdf
% * https://cogcomp.seas.upenn.edu/papers/DeutschDrRo22b.pdf: "Therefore, we recommend that reference-free metrics should be used as diagnostic tools for analyzing and understanding model behavior instead of measures of how well models perform a task, in which the goal is to achieve as high of a score as possible."
As there no ground-truth references exist for the adversarial example task, prior work has used a number of reference-free metrics to assess each method \citep[e.g.][]{greybox, BAE}, and we also follow the same approach. 

The first metric we use is the label-flip rate: the proportion of attacked examples where the victim model changes its predicted label compared to the original example. The next three are measures of text quality and are only computed for candidates that flip the label. The first of these is an assessment of the semantic similarity between the original and adversarial examples, and the metric chosen is BERTScore F1 \citep{BERTScore}, using a multilingual BERT variant as the base model. The second is a proxy assessment of fluency, for which we use BARTScore \citep{BARTScore} which is a measurement of the likelihood of a given output in a chosen multilingual BART model. Finally, the third metric assesses if the generated text is in the same language as the original. We assess this with the confidence output of a language detection model\footnote{We use a model from the Lingua library (https://github.com/pemistahl/lingua-py)}%, which is claimed to be more accurate than the commonly used langdetect tool (https://github.com/Mimino666/langdetect).}. %, 

% Rather than treat these metrics as a direct measure of performance, which is akin to using one generative model to evaluate another \citep{deutsch-etal-2022-limitations}

Eventually, we have combined these metrics into an overall metric, with the aim to identify candidates that do not obviously violate any of the adversarial example principles. We call this metric the Validated Success Rate (\textit{VSR}) and define it as the proportion of examples that have successfully flipped the predicted label while also meeting minimum thresholds on all of the text quality metrics. To set the similarity threshold, we have used as a guide the similarity scores achieved from the  multilingual paraphraser before training, which trend around 0.65. We have hence set a threshold of 0.6 to allow the generative model some `wiggle room' in the text generation. For the language detection, we have empirically set the threshold to 0.5. For the fluency threshold, we have explored how two different choices (``tight'', -9, and ``loose'', -11) affect the overall VSR achieved by each method, as shown in Figure \ref{fig:combined-results}. 

% \textbf{Postprocessing.} Before metric calculation, each successful attack has been post-processed to begin with a capital letter, end with a period, and have no whitespace around the last punctuation character.

\section{Results}
\label{sec:results}

The results for each language and  for the two different fluency thresholds are shown in Figure \ref{fig:combined-results}. For the proposed approach, we only plot one VSR value per threshold as a dot in correspondence of query limit = 32, since we use a beam size of 32 and all the beam outputs are assessed against the victim model. For the baselines, we plot full curves with maximum number of queries in the range [16, 4096] (blue: mBAE; red: mCLARE). The larger values correspond to the looser fluency threshold.

The results show that the performance from the proposed  generative model has been remarkable. It has outperformed the baselines in the low-query range by large margins, and has remained reasonably competitive for larger numbers of queries. Of the two baselines, mBAE has reported a higher performance for lower numbers of queries, and mCLARE for higher numbers. The performance across languages has been fairly consistent throughout, with the exception of Arabic, for which all methods have struggled. The baselines did more poorly over the German datasets, but this discrepancy can be partially explained by the impact of the document length: the test sets for German are made of shorter documents, and the token replacements made by the baselines affect the fluency metric more significantly. %Similarly French, on the other hand, contained longer documents. 

A few, selected qualitative examples of the generated attacks are shown in Table \ref{tab:samples}. These examples show that the generative approach has been able to produce effective adversarial examples by applying richer, sentence-level transforms. By contrast, the baselines' transforms have been more limited, and often not of high linguistic quality. 

% \section{Discussion}
% \label{sec:ablations}

\subsection{Sensitivity Analysis}
The semantic similarity and language detection coefficients in Equation \ref{eqn:training_objective} (respectively, $\alpha_{s}$ and $\alpha_{l}$) control the intensity of their respective regularisers, and in turn impact the quality of the generated text. We have qualitatively explored the impact of removing these components from the proposed approach, with examples displayed in Tables \ref{tab:ablation_sim} and \ref{tab:ablation_lang}. These examples confirm that removing the similarity component reduces the  faithfulness to the original example, and that removing the language detection component allows the generative model to ``stray'' to generate examples containing text in other languages.

% and substantially impact the quality and diversity of the generated text, as shown in Figure \ref{fig:kl_diversity_coeff_graph}. Increasing the KL coefficient ties the trained model more strongly to the reference model, which in our implementation increases the attack quality at the expense of the label-flipping rate. Conversely, lower values of the diversity coefficient push the model towards samples that are less diverse, while higher values promote diversity per se. Empirically, we have found that that the label-flipping rate has tended to remain constant for a range of diversity values, but the overall text quality metrics have peaked for a value of 10. 

\subsection{Language Impact}
%Although the masked language model used in the baselines, mBERT, performs a little better on French than on German \citep{wu-dredze-2020-languages}, the effect is not large enough to explain the discrepancy, especially since the used mT5 variant (Base) does as well \citep{mT5}. 

All methods have struggled with Arabic, and this suggests that specialist adversarial algorithms may be required, such as  \citet{Alshemali2019}. As a plausible reason, Arabic is highly inflected and differs significantly in its morphological and syntactic structure from the other four languages, making it unlikely that the adversarial attacks could effectively exploit their cross-lingual knowledge. Moreover, in the case of the proposed approach the base model, mT5, was originally trained on the mC4 corpus \citep{mT5}, which contains far fewer Arabic tokens than for the other languages, and the same applies to the TaPaCo corpus used for the paraphraser training. In turn, mCLARE and mBAE use multilingual BERT as the base language model, which was also trained on less Arabic data than for the other languages and is known to perform poorly compared to monolingual Arabic BERT variants, such as AraBERT \citep{antoun2020arabert}. Therefore, adversarial performance in Arabic may be improved by adopting dedicated language models. In addition, the BERT tokeniser also introduces a significant amount of redundancy in tokens when tokenising Arabic text \citep{antoun2020arabert}, and using a specialist Arabic segmenter such as Farasa \citep{Farasa} could improve performance. 

% Finally,  Arabic is known to use code-switching frequently, most commonly integrating French or English words \citep{Ali2021ArabicCS} which may have affected the performance of the language detector module or the langdetect system.

\section{Conclusion}
\label{sec:conclusion}

In this paper, we have presented a white-box generative approach for attacking multilingual text classifiers. The training objective of the proposed approach is very flexible, and incorporates specialised component models to encourage fluency, semantic consistency, and language adherence. To connect the generative model to the component models we have leveraged vocabulary-mapping matrices that allow using models of any vocabulary and retaining the full differentiability of the training objective. Experimental results across two multilingual datasets have confirmed the effectiveness of the proposed approach, particularly in a low number of queries scenario or when richer, sentence-level transforms are desired. %We have also presented a qualitative analysis of the key parameters on the characteristics of the generated text. 
Future research in this area might aim to adapt this framework to better target under-resourced or linguistically diverse languages, or to attack models for other multilingual NLP tasks, such as machine translation or cross-lingual classification. 

% \section{Ethical Considerations}

% \bibliography{custom}
% \bibliographystyle{acl_natbib}

% \clearpage

\appendix

\section{Training Details}
\label{sec:hyperparameters}
To train the generative model to paraphrase, we have used the TaPaCo dataset, using the splits for the chosen five languages.  After filtering out examples longer than our maximum token limit of 32 tokens, we had the following rough number of examples per language: ar: 6.5k, de: 125k, en: 158k, es: 85k, fr: 117k.  We then preprocessed these examples to create a corpus of ordered paraphrase pairs (i.e. so that (p1,p2) and (p2,p1) would be two separate paraphrase pairs). We then oversampled or undersampled the pairs to obtain 100k paraphrase pairs per language, and collated them to form our final training dataset.
For the MARC and TSM datasets, we have removed all training examples that the victim model classified incorrectly prior to the training procedure, as they could be said to be adversarial already. We have also removed examples longer than 32 tokens, since the paraphrase model was not trained for sequences in this range.
For the MARC dataset, we have categorised 1 and 2 star reviews as negative, 4 and 5 star reviews as positive, and discarded the 3 star reviews. Because this is a large dataset we selected a language-balanced subset to use as the training, validation and test sets, choosing 5k samples for the training set and 1k samples each for the validation and test sets. For the TSM dataset, the approximate sizes are in Table \ref{table:TSM_splits}.

\begin{table}[h]
\centering
\footnotesize
\begin{tabular}{l l}
\hline
\textbf{Lang} & \textbf{Trn / Val  / Test} \\
\hline
ar & 0.6k / 0.1k / 0.2k \\
de & 1.2k / 0.2k / 0.4k \\
en & 0.6k / 0.1k / 0.3k \\
es & 1.0k / 0.1k / 0.3k \\
fr & 0.9k / 0.1k / 0.4k \\
all & 4.3k / 0.6k / 1.6k \\
\hline
\end{tabular}
\caption{Approximate training/validation/test splits of the TSM dataset by language.}
\label{table:TSM_splits}
\end{table}
The hyperparameters used for our experiments are listed in Table \ref{tab:hyperparameters}. As optimiser, we have used Adafactor\footnote{We set the following arguments: scale\_parameter=False, relative\_step=False, warmup\_init=False)} with learning rate set to 0.0001. The models have been trained on a single NVIDIA A40 GPU with 48 GB RAM.

During training, we have bucketed examples with similar length to form training batches, and have passed the batches into the dataloader in random order.%  The details of all models used in the training objective are given in Table \ref{tab:models}. 

To select the values for the $\alpha$ parameters, we have first identified plausible ranges by initially performing a random search over the parameter space. We have then tuned the parameters manually, using subjective judgement based on our assessment of the quality of the generated text and the validation set performance. We have repeated the process for each dataset, as it is unlikely that a single set of parameters would be ideal for all. The $\beta$ coefficients have also been set manually, through an iterative process of examination of preliminary results and generated text samples, adjustment, and repetition, and a subjective examination of what constitutes an ``acceptable'' minimum threshold. The $\gamma$ constants in Section \ref{sec:validation}, which determine when to stop training, were also set in a similar way. The $\tau$ temperature parameter for the Gumbel-softmax sampling was set to 1.1 and we have not experimented with other values. 

\begin{table}[!ht]
\centering
\footnotesize
\begin{tabular}{@{}ll@{}}
\toprule
\textbf{Hyperparameter}                         & \textbf{Value}     \\ \midrule
\textit{General}                                &                    \\ \midrule
Optimisation algorithm                                              & Adafactor  \\
Learning rate                                              & $1 \times 10^{-5}$ \\
Weight decay                                              & 0 \\
Batch size   (train)                           & 5                \\
Max original length                         & 32                 \\
Precision & fp32 \\ 
% Seeds & 1114-1116 \\ 
\midrule
\textit{Coefficients (Equations \ref {eqn:training_objective} and  \ref{eqn:loss_fn}})                    &   \textit{MARC | TSM dataset}      \\ \midrule
Victim ($\alpha_v$) & 15 | 15 \\
Similarity ($\alpha_s$) & 20 |  25 \\
Language consistency ($\alpha_l$) & 20 | 20 \\
KL ($\alpha_{KL}$)  & 1 | 4 \\
Diversity ($\alpha_{d}$) & 0.1 | 10 \\
Victim threshold ($\beta_v$) & $1 - 1/c$ \\ 
Similarity threshold ($\beta_s$) & 0.75 \\ 
Language consistency threshold ($\beta_l$) & 0.0 \\ 
KL threshold ($\beta_{KL}$) & 2 \\ 
Gumbel-softmax temperature $\tau$  & 1.1 \\ 
\# Gumbel samples                     & 5 \\
\midrule
\textit{Early-stopping criteria (Section \ref{sec:validation})}                    &         \\ \midrule
Victim score  ($\gamma_v$) & 7 \\ 
Similarity score ($\gamma_s$) & 0.75 \\ 
Language consistency score ($\gamma_l$) & 1 \\ 
KL score ($\gamma_{KL}$) & 0.6 \\ 
Validation frequency  & Every 24 batches \\
Patience  & 12  \\
\midrule
\textit{Test-set generation}                    &         \\ \midrule
Batch size   (eval)                           & 5                 \\
\# generated sequences ($n$) & 32 \\
\# beams  & 32 \\
\# beam groups & 16 \\
Diversity penalty & 1 \\
Top-p & 0.98 \\ 
Temperature & 1 \\ 
Min generated length                       & $\max(0, l-2-\text{floor}(l/4))$                  \\
Max generated length                       & $l+2$                  \\
\bottomrule
\end{tabular}%

\caption{Hyperparameters.  $c$ is the number of classes in the dataset and $l$ is the batch length of generated text during evaluation (after padding).}
\label{tab:hyperparameters}
\end{table}

\section{Limitations}
One limitation of the proposed approach is that the experiments have only been conducted on datasets consisting of relatively short examples. However, since longer texts can always be subdivided, this limitation may not be very significant. 
A second limitation is that the approach has only been tested on five languages, and we do not know how it would perform in more diverse languages, such as Japanese or Chinese. A third limitation is that the method requires use of a GPU with a substantial amount of memory, in order to be able to store all the component models during training. The use of parameter-efficient fine-tuning methods, such as LoRA \citep{LoRA} may be able to mitigate this issue. 
% A limitation of the proposed approach is its substantial requirement for computing and hardware resources, as gradients need to be computed across multiple models. A GPU with at least 24 GB of memory is likely the minimum computing requirement, and for the utilization of larger models, correspondingly more memory is necessary. This limitation will lessen in the future as GPUs with larger memory will become routinely available, and as advancements in model distillation techniques continue. Another limitation is that the approach has been solely tested with a paraphrase model as the generator, and over datasets comprised of relatively short sentences. The generalizability of the method to other models and text styles remains then an open question. However, the possibility of subdividing larger blocks of text suggests that this may not be a significant limitation in practice.

\bibliography{aaai24}

\end{document}

%% file: examples_table.tex
\newcolumntype{L}[1]{>{\raggedright\let\newline\\\arraybackslash\hspace{0pt}}m{#1}}
\begin{table*}[ht]
    % \resizebox{\textwidth}{!}
    {%
    \setlength{\tabcolsep}{2pt}
    \footnotesize
        \begin{tabular}{lllL{2cm}llll}
            \toprule 
            \textbf{Dataset} &&&&& \textbf{Label} \\
            \midrule
            % \multirow{4}{*}{\bfseries HS} &
            % Orig & \multicolumn{3}{l}{Get two birds stoned at one time.}& Neither& \\
            % \cmidrule{2-6} &
            % mCLARE & \multicolumn{3}{l}{Get two chicks stoned at one time.}& Offensive& \\
            % \cmidrule{2-6} &
            % mBAE & \multicolumn{3}{l}{-}& Unsuccessful& \\
            % \cmidrule{2-6} &
            % Ours & \multicolumn{3}{l}{At the same time get two birds stoned.}& Offensive& \\
            % \midrule
 
            \multirow{8}{*}{\bfseries MARC} &
            Orig & \multicolumn{3}{l}{I love that it keeps the wine cold.}& Positive& \\
            \cmidrule{2-6} &
            mCLARE & \multicolumn{3}{l}{I \textcolor{blue}{not} love that it keeps the wine cold.}& Negative& \\
            % \cmidrule{3-6}
            % \cmidrule{3-6} 
			&	& \textit{Flu}: -7.11 \quad \textit{Lang}: 0.84 \quad \textit{Sim}:  0.96 \quad \\
            \cmidrule{2-6} &
            mBAE & \multicolumn{3}{l}{I \textcolor{blue}{doubt} that it keeps the wine cold.}& Negative& \\
			&	& \textit{Flu}: -7.60 \quad \textit{Lang}: 0.85 \quad \textit{Sim}:  0.92 \quad \\
            \cmidrule{2-6} &
            Ours & \multicolumn{3}{l}{In my opinion, it keeps the wine cold!}& Negative& \\
			&	& \textit{Flu}: -7.63 \quad \textit{Lang}: 0.78 \quad \textit{Sim}:  0.82 \quad \\
        
            \midrule
            \midrule
            \multirow{8}{*}{\bfseries MARC} &
            Orig & \multicolumn{3}{l}{Une batterie neuve qui fait à peine mieux que mon ancienne batterie. Décevant. }& Negative& \\
            \cmidrule{2-6} &
            mCLARE & \multicolumn{3}{l}{Une batterie neuve qui fait à peine mieux que mon ancienne batterie. \textcolor{blue}{Ravel}.}& Positive& \\
            % \cmidrule{3-6}
            % \cmidrule{3-6} 
			&	& \textit{Flu}: -5.24 \quad \textit{Lang}: 0.97 \quad \textit{Sim}:  0.97 \quad \\
            \cmidrule{2-6} &
            mBAE & \multicolumn{3}{l}{Une batterie neuve qui fait \textcolor{blue}{vraiment bon} mieux que mon ancienne batterie. Décevant.}& Positive& \\
			&	& \textit{Flu}: -4.89 \quad \textit{Lang}: 0.96 \quad \textit{Sim}:  0.99 \quad \\
            \cmidrule{2-6} &
            Ours & \multicolumn{3}{l}{C'est une batterie neuve qui est bien mieux que mon ancienne batterie.}& Positive& \\
			&	& \textit{Flu}: -6.74 \quad \textit{Lang}: 0.99 \quad \textit{Sim}:  0.75 \quad \\
        
            \midrule
            \midrule
            \multirow{8}{*}{\bfseries TSM} &
            Orig & \multicolumn{3}{l}{Heute Abend dann noch Haare färben und Sport machen :D}& Positive& \\
            \cmidrule{2-6} &
            mCLARE & \multicolumn{3}{l}{Heute Abend dann noch Haare färben und Sport machen \textcolor{blue}{:Roman} }& Neutral& \\
			&	& \textit{Flu}: -8.19 \quad \textit{Lang}: 0.88 \quad \textit{Sim}:  0.99 \quad \\
            \cmidrule{2-6} &
            mBAE & \multicolumn{3}{l}{Heute Abend dann noch Haare färben und Sport machen \textcolor{blue}{:Roman}}& Neutral & \\
			&	& \textit{Flu}: -8.19 \quad \textit{Lang}: 0.88 \quad \textit{Sim}:  0.99 \quad \\
            \cmidrule{2-6} &
            Ours & \multicolumn{3}{l}{Was hältst du davon, heute Abend noch Haare färben und Sport}& Neutral& \\
			&	& \textit{Flu}: -6.58 \quad \textit{Lang}: 0.97 \quad \textit{Sim}:  0.71 \quad \\
            \midrule
            \midrule
   %          \multirow{4}{*}{\bfseries TSM} &
   %          Orig & \multicolumn{3}{l}{انا ايه اللي خلاني اتعلم اللغة الفرنسية و انا مش هسافر فرنسا اصلاً}& Negative& \\
   %          \cmidrule{2-6} &
   %          mCLARE & \multicolumn{3}{l}{انا ايه اللي خلاني اتعلم اللغة الفرنسية و انا مش هسافر فرنسا [UNK]}& Positive& \\
			% &	& \textit{Flu}: 23.99 \quad \textit{Lang}: 20.00 \quad \textit{Sim}:  20.77 \quad \\
   %          \cmidrule{2-6} &
   %          mBAE & \multicolumn{3}{l}{[UNK] ايه اللي خلاني اتعلم اللغة الفرنسية و انا مش هسافر فرنسا اصلاً }& Positive& \\
			% &	& \textit{Flu}: 23.99 \quad \textit{Lang}: 20.00 \quad \textit{Sim}:  20.77 \quad \\
   %          \cmidrule{2-6} &
   %          Ours & \multicolumn{3}{l}{انتا سأتعلم اللغة الفرنسية وأنا مشغولة فرنسا فقط!}& Positive& \\
			% &	& \textit{Flu}: 23.99 \quad \textit{Lang}: 20.00 \quad \textit{Sim}:  20.77 \quad \\
   %          \midrule
            % \multirow{2}{*}{\bfseries FP} &
            % Orig & \multicolumn{3}{l}{In addition , the company will reduce a maximum of ten jobs.}& Negative& \\
            % \cmidrule{2-6} &
            % Adv & \multicolumn{3}{l}{It has announced it has a maximum of ten job-separation measures. It has announced it has}& Neutral& \\
            % \midrule
            % \multirow{4}{*}{\bfseries Emotion} &
            % Orig & \multicolumn{3}{l}{i find myself feeling anxious and unsure}& Fear& \\
            % \cmidrule{2-6} &
            % mCLARE & \multicolumn{3}{l}{i find myself looking curious and unsure}& Joy& \\
            % \cmidrule{2-6} &
            % mBAE & \multicolumn{3}{l}{-}& Unsuccessful& \\
            % \cmidrule{2-6} &
            % Adv & \multicolumn{3}{l}{anxiety and a lack of confidence}& Joy& \\
            % \midrule
        \end{tabular}
    }
    \caption{Examples of successful adversarial attacks generated by the various methods for three languages. The token changes of the baselines are highlighted in blue, and the values of the fluency, language consistency, and semantic similarity are reported below. The examples show that our method has been able to create richer, coherent adversarial examples than the baselines, which are limited to modifying individual tokens. To generate successful examples, the  baselines often employ ``tricks'' such as inserting negatives (first example), forming ungrammatical sentences (e.g. \textit{bon mieux} for mBAE in the second example, or the misformed emoji \textit{:Roman} in the third) and changing the base language (\textit{Ravel} in the second example). }
    \label{tab:samples}
\end{table*}

% “ funny valentine “ is about learning what it takes to find true love 
%  “ funny valentine “ is about inside what it takes to find true love 
% I'm learning what it takes to find true love 

%   what is the atomic weight of silver?
%  what is the atomic composition of silver? 
%  I'm sorry but that's the atomic weight of silver

%  get two birds stoned at one time 
%  get two chicks stoned at one time . 
% -
% At the same time get two birds stoned.

% i find myself feeling anxious and unsure  
% i find myself looking curious and unsure 
% --

% suffers from unlikable characters and a self-conscious sense of its own quirky hipness.
% . from unlikable characters and a self-conscious sense of its own quirky hipness.
% signs of unlikable characters and a self-consciousness sense of its own quirky hipness.
% it is characterized by characters who are unlikable and it has a sense of hipness that is self-conscious.

%% file: ablation_sim_table.tex
% \begin{table}[h]
% \centering
% \begin{tabular}{lll}
% \toprule
% \textbf{Method} & \textbf{Text} & \textbf{Label} \\
% \midrule
% Orig & These raw cashews are delicious and fresh. & Positive \\
% Ours & I'll like this raw cashew. & Negative \\
% Ours (no sim) & I'm going to eat these raw cashiers. & Negative \\
% \midrule
% Orig & I love that it keeps the wine cold. & Positive \\
% Ours &In my opinion, wine stays cold. & Negative \\
% Ours (no sim) &  The wine is delicious. I'm going to like & Negative \\
% \bottomrule
% \end{tabular}
% \caption{Enhanced Presentation of Text Classification Results}
% \label{tab:classification_results}
% \end{table}

\begin{table}[h]
\centering
\footnotesize
\begin{tabularx}{\columnwidth}{l>{\raggedright\arraybackslash}Xl}
\toprule
\textbf{Mode} & \textbf{Text} & \textbf{Label} \\
\midrule
Orig & These raw cashews are delicious and fresh. & Pos \\
$+$Sim & I'll like this raw cashew. & Neg \\
$-$Sim & I'm going to eat these raw \textcolor{red}{cashiers}. & Neg \\
\midrule
Orig & I love that it keeps the wine cold. & Pos \\
$+$Sim & In my opinion, it keeps the wine cold! & Neg \\
$-$Sim & The wine is \textcolor{red}{delicious}. I'm going to like & Neg \\
\bottomrule
\end{tabularx}
\caption{Illustrative English examples from training with and without the semantic similarity component. In the first example, \textit{cashew} is changed to \textit{cashier} --- a simple change of one subword token that loses the sentence meaning. In the second example, the information about the temperature of the wine is lost.}
\label{tab:ablation_sim}
\end{table}
\vspace{-6pt}

%% file: ablation_lang_table.tex
\begin{table}[h]
\centering
\footnotesize
\begin{tabularx}{\columnwidth}{l >{\raggedright\arraybackslash}X l}
\toprule
\textbf{Mode} & \textbf{Text} & \textbf{Label} \\
\midrule
Orig & Super, hilft schnell & Pos \\
$+$LD & Irgendetwas, das ist schnell. & Neg \\
$-$LD & Solch ein \textcolor{red}{ayuda}! & Neg \\
\midrule
Orig & Ces sachets sentent le plastique. & Neg \\
$+$LD & Ces sachets sont pour le plastique! & Pos \\
$-$LD & Ces sachets\textcolor{red}{gefühlt} le plastique. & Pos \\
\bottomrule
\end{tabularx}
\caption{Training without the language detection component often leads to foreign words in the generated text. In the first example, the model has directly translated the German \textit{hilft} (to help) to the Spanish word \textit{ayuda}. Similarly, in the second example the French  \textit{sentent} (to smell) is translated into the German \textit{gefühlt}. }
\label{tab:ablation_lang}
\end{table}